\documentclass{article}

\usepackage{arxiv}

\usepackage[utf8]{inputenc} % allow utf-8 input
\usepackage[T1]{fontenc}    % use 8-bit T1 fonts
\usepackage{hyperref}       % hyperlinks
\usepackage{url}            % simple URL typesetting
\usepackage{booktabs}       % professional-quality tables
\usepackage{amsfonts}       % blackboard math symbols
\usepackage{nicefrac}       % compact symbols for 1/2, etc.
\usepackage{microtype}      % microtypography
\usepackage{lipsum}		% Can be removed after putting your text content
\usepackage{graphicx}
\usepackage{natbib}
\usepackage{doi}
\usepackage{float}

\usepackage[frozencache,cachedir=minted-cache]{minted} 

\title{The CAST package for training and assessment of spatial prediction models in R}

\author{ 
\hspace{1mm}Hanna~Meyer \\
	Institute of Landscape Ecology,\\
	University of Münster\\
	Münster, Germany\\
	\texttt{hanna.meyer@uni-muenster.de} \\
	\And
\hspace{1mm}Marvin~Ludwig \\
	Institute of Landscape Ecology,\\
	University of Münster\\
	Münster, Germany\\
	\texttt{marvin.ludwig@uni-muenster.de} \\
 \And
 \hspace{1mm}Carles~Milà \\
    ISGlobal, Barcelona Institute for Global Health \& \\
    Universitat Pompeu Fabra,\\    Barcelona, Spain\\
    \texttt{carles.mila@isglobal.org} \\
  \And
 \hspace{1mm}Jan~Linnenbrink \\
	Institute of Landscape Ecology,\\
	University of Münster\\
	Münster, Germany\\
	\texttt{jan.linnenbrink@uni-muenster.de} \\
  \And
 \hspace{1mm}Fabian~Schumacher \\
	Institute of Landscape Ecology,\\
	University of Münster\\
	Münster, Germany\\
	\texttt{fabian.schumacher@uni-muenster.de} \\
 }

\renewcommand{\shorttitle}{R package CAST}

\hypersetup{
pdftitle={Introduction to CAST},
pdfauthor={Hanna Meyer, Marvin Ludwig, Carles Mila, Jan Linnenbrink, Fabian Schumacher}
}

\begin{document}
\maketitle

\begin{abstract}

One key task in environmental science is to map environmental variables continuously in space or even in space and time. Machine learning algorithms are frequently used to learn from
local field observations to make spatial predictions by estimating the value of the variable of interest in places where it has not been measured. 
However, the application of machine learning strategies for spatial mapping involves additional challenges compared to "non-spatial" prediction tasks that often originate from spatial autocorrelation and from training data that are not independent and identically
distributed. 

In the past few years, we developed a number of methods to support the application of machine learning for spatial data which involves the development of suitable cross-validation strategies for performance assessment and model selection, spatial feature selection, and methods to assess the area of applicability of the trained models. The intention of the CAST package is to support the application of machine learning strategies for predictive mapping by implementing such methods and making them available for easy integration into modelling workflows.

Here we introduce the CAST package and its core functionalities. At the case study of mapping plant species richness, we will go through the different steps of the modelling workflow and show how CAST can be used to support more reliable spatial predictions.

\end{abstract}


\section{Introduction}
One key task in environmental science is to map environmental variables continuously in space or even in space and time as a baseline to inform decision-making in various applications, such as agriculture, land-use planning, or natural resource management; or to study ecological research questions based on spatial patterns. However, most ecological variables are only available as point data, e.g. from field surveys. Modelling approaches are hence required to move from local field observations to continuous maps of ecological variables by estimating the value of the variable of interest in places where it has not been measured. In recent years, machine learning methods have become a popular tool to learn patterns in nonlinear and complex systems. These methods have been applied to map various ecological variables, even on a global scale, such as soil properties \citep{Hengl2017, Lembrechts2022}, plant traits \citep{Moreno-Martinez2018}, or occurrence and abundance of plant \citep{Sabatini2022, Cai2023} and animal species \citep[e.g. soil nematodes][]{VanDenHoogen2019}. 

\begin{figure}[ht!]
\includegraphics[width=\columnwidth]{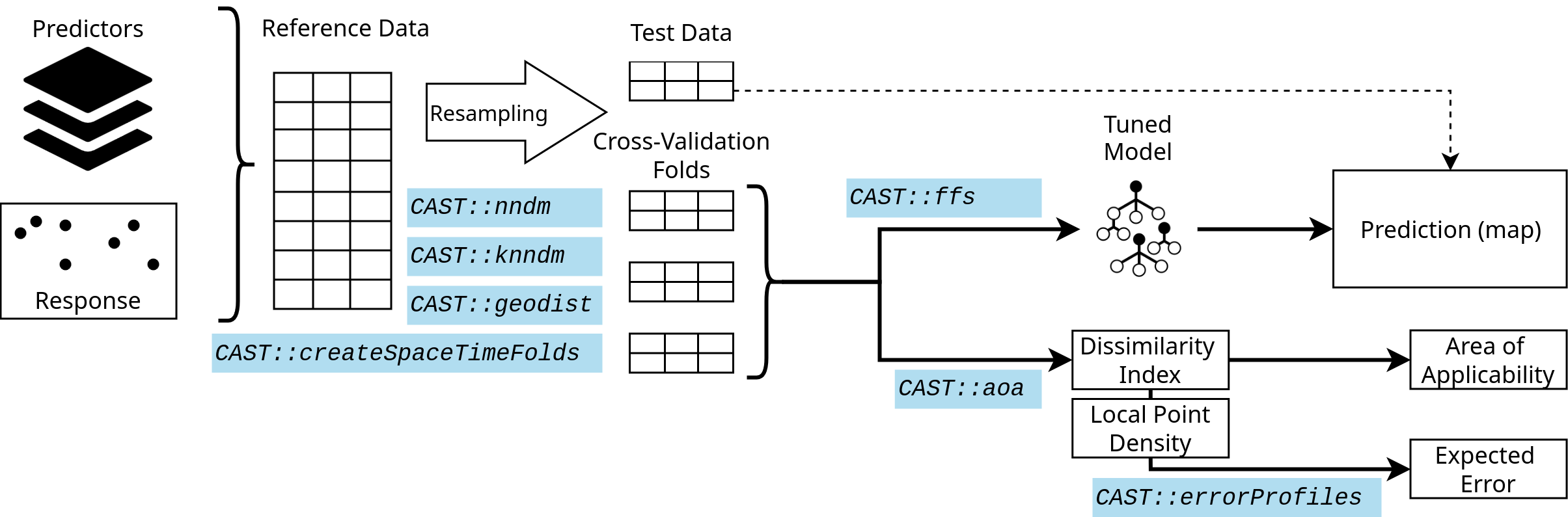}
\caption{A very simple workflow for a spatial prediction mapping workflow, indicating which function in CAST can be used in the different steps to support the spatial prediction.}
\label{fig:WorkflowAndCAST}
\end{figure}

These studies all follow a very similar workflow (Figure \ref{fig:WorkflowAndCAST}).  First, reference data of the response variable are collected, both for model training as well as for model validation. These are often derived from field surveys (e.g. soil sampling, plot-based vegetation surveys, measurements from weather stations), where often data from various studies are compiled in larger databases such as sPlotOpen \citep{Sabatini2021_splotopen} for vegetation surveys or TRY \citep{Kattge2020} for plant trait data. Second, a set of potential predictor variables (or "features") that are assumed to be drivers of the response variable is compiled. These variables are available for the entire area of interest and often originate from remote sensing data sources such as spectral channels from optical sensors. These might also be variables describing the climate \citep[e.g. WorldClim,][]{Fick2017} or terrain properties. Third, a supervised model is then trained to learn the relationships between the response variable and the corresponding predictor values. Among the many available algorithms, random forest is one the most popular machine learning algorithm for spatial modelling tasks. The model training process typically involves several steps including hyperparameter tuning and feature selection, where typically cross-validation is used to optimize the parameters or to select the variables used in the model. Fourth, the trained model is then applied to each pixel of the predictor set to generate spatial maps of the environmental response variable of interest. Finally, the assessment of the map quality as well as uncertainty assessment is performed as an essential part of the modelling process to inform the producer and user of the reliability of the generated product.

R is a very popular software environment and programming language for spatial predictive modeling and all its involved steps. There is a large number of machine learning algorithms implemented in various packages, which can also be accessed via wrapper packages such as caret \citep{Kuhn2008}, tidymodels \citep{Kuhn2020} or mlr3 \citep{Lang2019}. All of these three wrapper packages simplify and standardize the machine learning workflow in R, providing users with a consistent interface for model training, prediction and evaluation.
However, the application of machine learning strategies for spatial mapping involves additional challenges compared to other "non-spatial" prediction tasks that often originate from spatial autocorrelation and from training data that are not independent and identically distributed (i.i.d) \citep[see e.g.][]{Meyer2022}. In the past few years, we developed a number of methods to support the application of machine learning for spatial data which involves the development of suitable cross-validation strategies for performance assessment and model selection \citep{Mila2022, LinnenbrinkInReview}, spatial feature selection \citep{Meyer2018,Meyer2019}, and methods to assess the area of applicability of the trained models \citep{Meyer2021}.

The intention of the CAST package is to support the application of machine learning strategies for predictive spatial mapping by implementing such methods and making them available for easy integration into modelling workflows. CAST stands for "Caret Applications for Spatio-Temporal models" which indicates that the package was initially (in 2018) developed as a wrapper around caret \citep{Kuhn2008}. However, with more users shifting towards mlr3 or tidymodels, we intend to make the functionality available beyond the caret package by making it usable with other wrapper packages as well.

With this chapter, we want to show how the CAST package can be used to close important gaps in the machine learning-based spatial mapping workflow. For each of the main CAST functionalities (Figure \ref{fig:WorkflowAndCAST}), we will first explain the concept and idea behind the methods and then show its practical application with an example dataset of plant species diversity in South America.

\section{Functionality of the CAST package}

Alongside the implementation in CAST, all methods are published in scientific papers. Currently, the main features of CAST are:

\begin{itemize}
    \item The cross-validation strategy "Nearest Neighbor Distance Matching" (NNDM) cross-validation \citep{Mila2022} and its k-fold variant kNNDM cross-validation \citep{LinnenbrinkInReview}, which are designed to overcome limitations of existing cross-validation strategies to provide estimates of map accuracy. The package further implements additional support for user-defined spatial or spatio-temporal assignment of cross-validation folds as explained in \citet{Meyer2018}. See \texttt{CAST::nndm}; \texttt{CAST::knndm}; \texttt{CAST::createSpaceTimeFolds}.
    \item Visualization methods as used in \citet{Mila2022, Meyer2022, Ludwig2023} to inspect the representativeness of the cross-validation folds for a spatial prediction task. See \texttt{CAST::geodist}.
    \item Methods for spatial feature selection that are designed to select predictor variables that are most suitable for spatial mapping and minimize the risk of overfitting \citep{Meyer2018, Meyer2019}. Application for large scale spatial prediction tasks are shown in \citet{Ludwig2023}. See \texttt{CAST::ffs}.
    \item A method for the assessment of the area of applicability \citep{Meyer2021} that is defined as the area where the model was enabled to learn about relationships and where the estimated cross-validation performance may hold. See \texttt{CAST::aoa}.
    \item A method to estimate prediction uncertainties based on distances and data point densities in the predictor space \citep{Meyer2021}. See \texttt{CAST::errorProfiles}.
\end{itemize}

The package is available on CRAN \url{https://CRAN.R-project.org/package=CAST}, the developer version is on Github \url{https://github.com/HannaMeyer/CAST}. Here we used version 1.0.0. The package is supported by several vignettes that explain the functionality of each of its components. They can be found on the documentation page of the package, on \href{hannameyer.github.io/CAST/}{https://hannameyer.github.io/CAST/}.

%%%%%%%%%%%%%%%%%%%%%%%%%%%%%%%%%%%%%%%%%%%%%%%%%%%%%%%
\section{Example data and modelling task}
\begin{figure}[H]
\includegraphics[width=\textwidth]{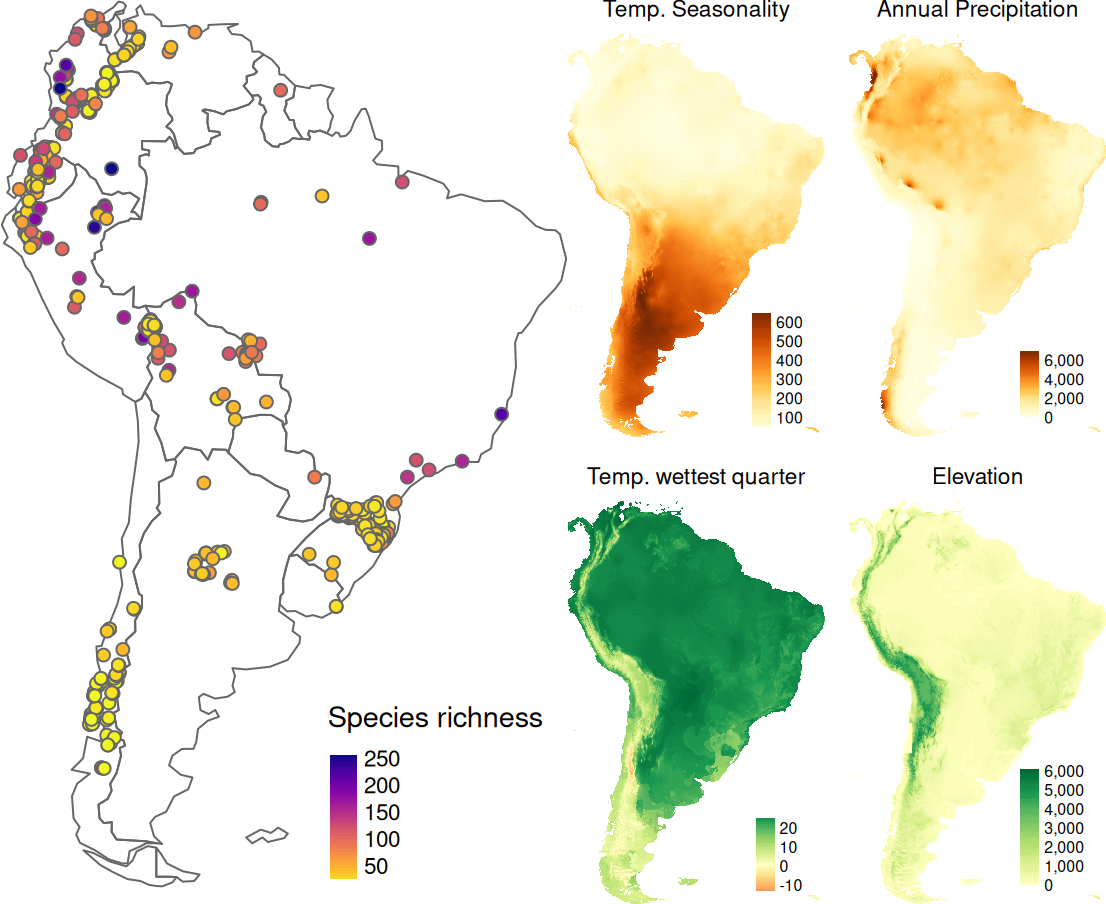}
\caption{Location of the reference data from sPlotOpen and example predictor variables (an excerpt of \emph{predictors\_raster}) for the desired prediction domain.}
\label{fig:predictorsResponse}
\end{figure}

%%%%%%%%%%%%%%%%%%%%%%%%%%%%%%%%%%%%%%%%%%%%%%%%%%%%%%%
We assume that the reader of this chapter is already familiar with basic principles of applied predictive modelling \citep[if not, we recommend][as a starting point]{Kuhn2013, james2013}. For demonstration of the CAST functionalities, we will go through a typical spatial prediction task, which aims here at producing a spatially-continuous map of plant species richness for South America. As reference data, we use plant species richness data from plot-based vegetation surveys that are compiled in the sPlotOpen database described in \citet{Sabatini2021_splotopen} \citep[datasets used:][]{Pauchard2013, Peyre2015, Lopez-Gonzalez2011, Vibrans2020}. We will use WorldClim climatic variables \citep{Fick2017} and elevation \citep{Jarvis2008} as predictors, assuming that they are relevant drivers of species richness. The data are visualized in Figure \ref{fig:predictorsResponse}. Random forest is used here as machine learning algorithm.
Please note that this example is build here for technical demonstration purposes and should not be regarded as a reference for mapping plant species richness. 

The code and data used for the example described in this chapter is available at \href{loek-rs.github.io/CAST4ecology}{https://loek-rs.github.io/CAST4ecology}. Therein, the predictor variables for entire South America are available as multi-band GeoTIFF (here referred to as \emph{predictors\_raster}). The reference data from sPlotOpen are available as sf object as part of the CAST package (called here \emph{training\_data}) and contain the predictor information as attributes (called here \emph{predictor\_names}), as well as the species richness information (called here \emph{response\_name}).
For handling of raster data, we will rely on the terra package \citep{terra}. For the vector data, sf will be used \citep{pebesma_sf1, pebesma_sf2}.

\subsection{A first simple prediction model}
As a first example we will see how we can train a "default" model without consideration of the challenges of spatial prediction that will be discussed next. When using the caret package, we can use the \texttt{train} function, which takes the predictors and response from the training data as well as the information of the algorithm to be used as arguments \citep[here: random forest with 100 trees from the ranger package, ][]{Wright2017}. Default random cross-validation is used for hyperparameter tuning and assessment. Once the model is trained we can use it to make spatial predictions (see code below).

\vspace{0.3cm}
\begin{minted}[fontsize=\small]{R}
# train model:
rfmodel_rcv <- caret::train(x = training_data[,predictor_names],
                            y = training_data[,response_name],
                            method = "ranger",
                            trControl = trainControl(method = "cv"))
# get validation statistics and predict:
CAST::global_validation(rfmodel_rcv)
prediction <- terra::predict(predictors_raster, rfmodel_rcv, na.rm = TRUE)
\end{minted}

\begin{figure}[H]
\centering
\includegraphics[width=.5\textwidth]{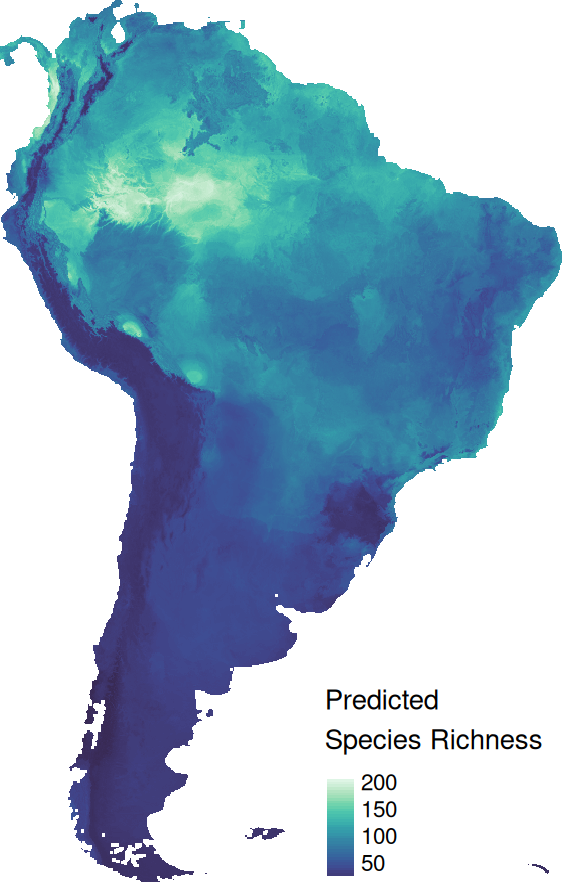}
\caption{A first prediction of plant species richness in South America.}
\label{fig:firstprediction}
\end{figure}

In this first model we see that, technically, we can easily make predictions for the entire study area as shown in Figure \ref{fig:firstprediction} (and might even go beyond that). However, these predictions also raise many questions, e.g how much should we trust them? How could we improve the model? Can we really apply the model trained on so few and clustered reference locations to make predictions for entire South America? What are the uncertainties in doing so? In the next sections, we will see how we can use the CAST package to deal with these concerns.

%%%%%%%%%%%%%%%%%%%%%%%%%%%%%%%%%%%%%%%%%%%%%%%%%%%%%%%
\section{Cross-validation to estimate the map accuracy}
%%%%%%%%%%%%%%%%%%%%%%%%%%%%%%%%%%%%%%%%%%%%%%%%%%%%%%%
A map as produced by the workflow described above is only valuable when we have an accuracy estimate that measures how much (or little) we can trust it. For an unbiased estimation of the map accuracy, a probability sample of the entire prediction area is required \citep[e.g.][]{Stehman1999, Wadoux2021}. The problem is that this is rarely available in ecology \citep[see e.g.][]{Meyer2022} because reference data are often clustered in space while some areas are left completely unsampled. This is not only problematic for the map evaluation, but also for the modelling process itself since an estimation of the map accuracy is used during model tuning, i.e. to identify the best set of hyperparameters and/or predictor variables.
To account for the absence of an external probability sample used for testing the final predictions, the available reference data are usually split into 3 data sets: training data, validation data (used during model tuning) and test data (used to evaluate the final predictions). Data resampling into training and validation sets (so called folds) is usually done several times in a cross-validation procedure. Test data, on the other hand, are usually selected once beforehand and ideally according to the same resampling rules that are applied for defining the cross-validation folds. 
Here, we will focus on cross-validation; however, the same concepts and functions can be used for a prior split into training and test data.

\begin{figure}[H]
\centering
\includegraphics[width=\textwidth]{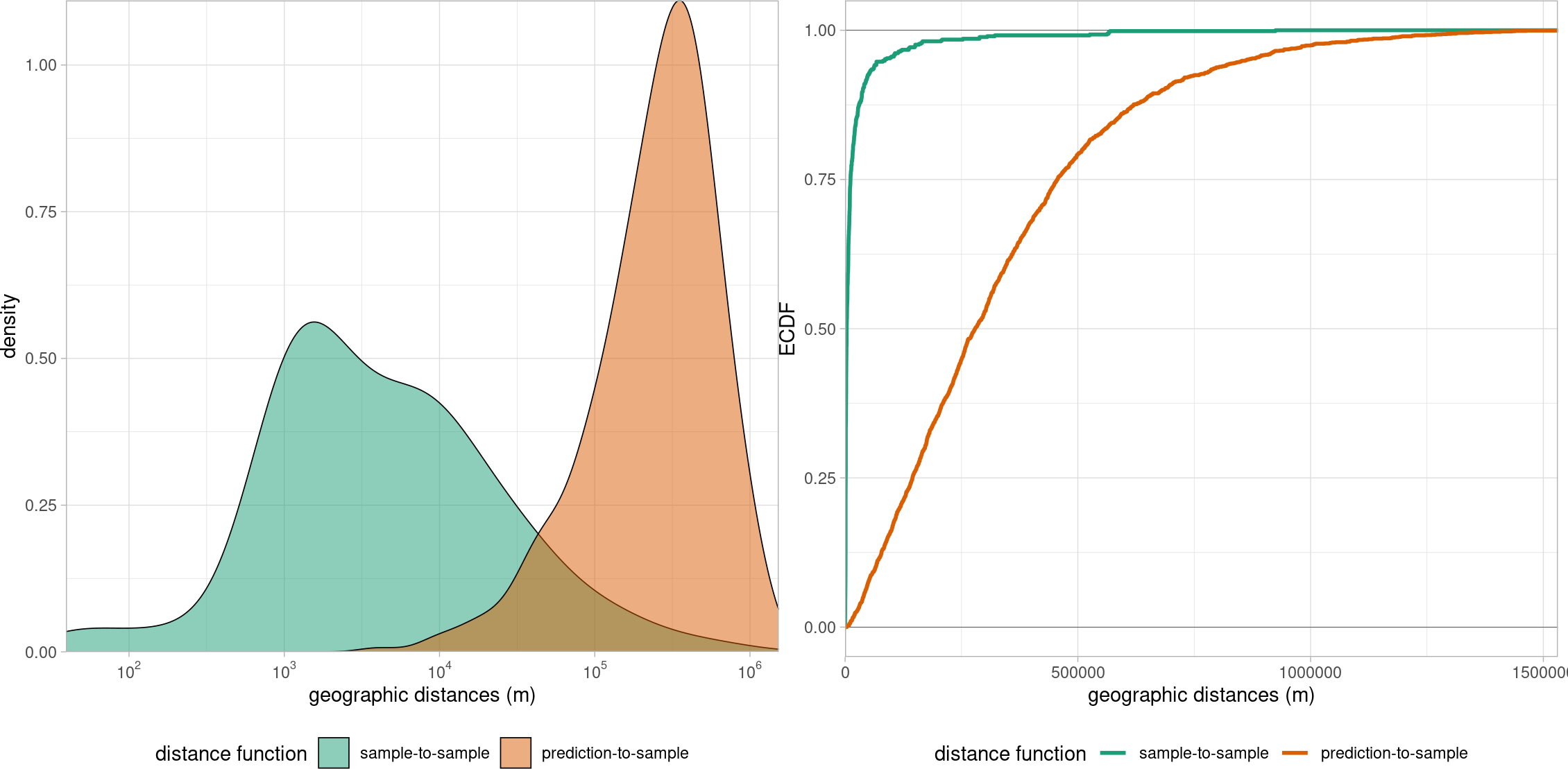}
\caption{Nearest neighbor distance distribution represented as density plot (left) and empirical cumulative distribution function plot (right). Both show that prediction requires an application of the model far beyond the clustered reference data. Aim of the cross-validation strategies implemented in CAST is to produce similar prediction-to-sample distances based on the available training data.}
\label{fig:geodist_nocv}
\end{figure}

Whether the (cross-)validation can give a reliable estimate of the map accuracy depends on the strategy we use to split the data. It has been reported several times that the commonly applied random cross-validation is not suitable for spatial data. It assumes data are independent and results in overoptimistic estimates when sampling data are clustered \citep{Wadoux2021, Mila2022}. Spatial cross-validation strategies that try to achieve independence between test and train data \citep[e.g.][]{Valavi2019,Roberts2017,Brenning2001}, however, may have shortcomings as well especially if training data are randomly or regularly distributed \citep{Wadoux2021, Mila2022}. To address this issue, in \citet{Meyer2022, Mila2022} we argue that the cross-validation strategy used for map accuracy estimation should be prediction-oriented; i.e. it should aim to produce similar predictive conditions to those found when using the model to predict a new area. In the case of spatial prediction models, the "predictive conditions" can refer to the geographical Nearest Neighbour Distances (NND) between prediction and training points. Figure \ref{fig:geodist_nocv} shows the challenge in our example: since our training samples are clustered within the prediction area, we see that NND between training points are generally much shorter than NND between prediction and training points. The latter is the situation the model faces during prediction (i.e. the model needs to be applied to areas that are far away from a reference data point, and thus we should try to resemble it during cross-validation to get a meaningful estimate of the map accuracy).
Note that \texttt{geodist} can be calculated in both, geographical space and feature space. Visualization allows for density plots (default, see Figure \ref{fig:geodist_nocv} left) or empirical cumulative distribution function plots (see Figure \ref{fig:geodist_nocv} right). 

\vspace{0.3cm}
\begin{minted}[fontsize=\small]{R}
# compare distances in geographic space:
geo_distance <- CAST::geodist(training_data, 
                              modeldomain=predictors_raster)
plot(geo_distance)
plot(geo_distance, stat="ecdf")
\end{minted}

With the objective of providing a prediction-oriented cross-validation strategy suitable for spatial prediction tasks, we proposed the Nearest Neighbor Distance Matching (NNDM) algorithm \citep{Mila2022}, a Leave-One-Out (LOO) cross-validation method that aims to create a distribution of NND between test and train data during cross-validation that matches the one found during prediction. Briefly, NNDM uses spatial point pattern concepts to exclude additional samples when performing a LOO cross-validation so that the Empirical Cumulative Distribution Function (ECDF, see Figure \ref{fig:geodist_nocv} right) of NND between test and train points during cross-validation matches the ECDF of NND between prediction and train points \citep[see][for more information and simulations evaluating its performance]{Mila2022}. Nonetheless, as a LOO method, it is computationally intensive and cannot be used with large datasets. Therefore, in \citet{LinnenbrinkInReview}, we further suggest a k-fold variant of this approach: the kNNDM algorithm.

kNNDM uses a clustering approach in which the reference samples are clustered in a varying number of groups, and then merged into the desired final number of \textit{k} folds. Among all the candidate groupings, the one that minimises the Wasserstein distance (absolute value integral between the two curves) between the ECDF of NND between test and train points during cross-validation and the ECDF of NND between prediction and train points, i.e. the one that offers the best match, is chosen. 

\begin{figure}
    \centering
    \includegraphics[width=\textwidth]{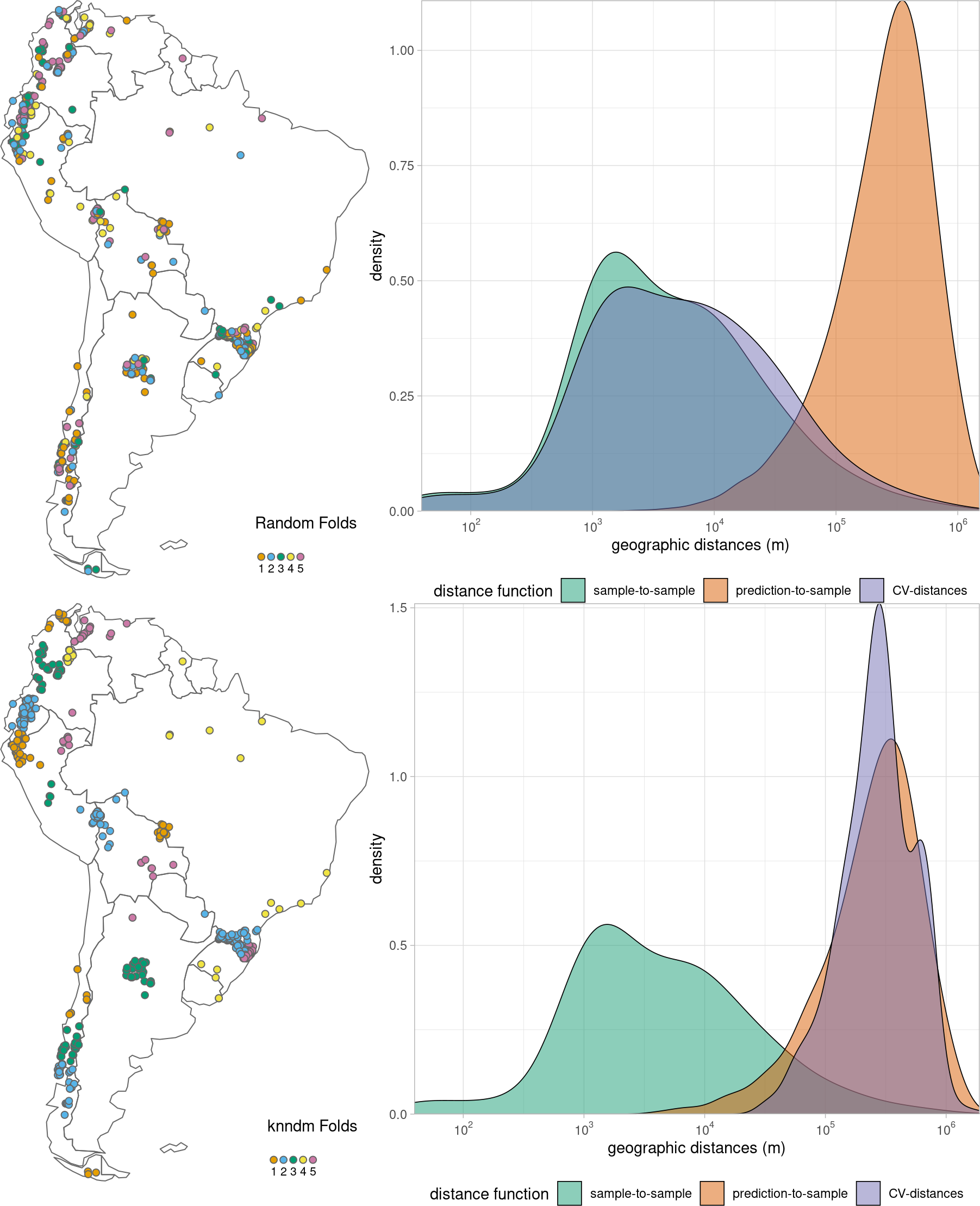}
    \caption{Comparison of cross-validation methods: random folds and their corresponding nearest neighbor distance distribution (top) as well as kNNDM folds and their nearest neighbor distance distribution (bottom)}
    \label{fig:cv_comparison}
\end{figure}

Going back to our example, if we just use a random k-fold cross-validation, the distribution of NND between cross-validation folds closely matches the distribution of NND between train points (Figure \ref{fig:cv_comparison} top). Thus, when using random k-fold cross-validation we are not testing the ability of the model to make predictions for all South America, but rather evaluating how good the model can reproduce the training data. We now use the kNNDM algorithm to assign the sample points to cross-validation folds in a way that resembles the prediction conditions. The resulting cross-validation split assigns the reference points to spatially distinct folds, thus extending nearest neighbour distances. This results in a better match between the NND during cross-validation and the NND between prediction and training points (Figure \ref{fig:cv_comparison}). 

\vspace{0.3cm}
\begin{minted}[fontsize=\small]{R}
# define cross-validation folds with knndm: 
knndm_folds <- CAST::knndm(tpoints = plots,
                           modeldomain = predictors_raster, k = 5)
\end{minted}

\vspace{0.3cm}
We can now use the resulting cross-validation folds to estimate the map accuracy and find the best hyperparameters during model tuning. The R-code is mostly the same as the one to train the default random forest model, we just need to specify the cross-validation folds as the index-parameter in caret's \texttt{trainControl} function:

\vspace{0.3cm}
\begin{minted}[fontsize=\small]{R}
# indicate the use kNNDM cross-validation folds:
tr_control <- caret::trainControl(method = "cv",
                                  index = knndm_folds$indx_train,
                                  savePredictions = TRUE)
# tune and train model:
rfmodel_knndmcv <- caret::train(x = training_data[,predictor_names],
                                y = training_data[,response_name],
                                method = "ranger",
                                trControl = tr_control)
# get validation statistics and predict:
CAST::global_validation(rfmodel_knndmcv)
prediction <- terra::predict(rfmodel_knndmcv, predictors_raster)
\end{minted}

\vspace{0.5cm}
Note that we also set the parameter \texttt{savePredictions = TRUE}. This allows us to stack the different cross-validation predictions afterwards to calculate the validation statistics using the function \texttt{CAST::global\_validation}. This is relevant because kNNDM matches the distribution of NND based on all the folds simultaneously and can produce folds with different number of samples, and thus the traditional approach of computing validation statistics in each fold and then averaging them would not produce the desired results (especially for validation statistics such as $R^2$). If we now compare the results of the random and the kNNDM cross-validation, we see that the random k-fold cross-validation results in much better validation metrics (Table \ref{tab:cv_results}). This is due to the fact that we tested the performance of the model holding out points that are very close to the training points, making it easy for the model to predict their values, which probably leads to overoptimism in the estimated map accuracy \citep{Mila2022, Wadoux2021}. The kNNDM cross-validation results in worse validation metrics, because the hold-out points are further away from the training points and, similar to what we will find when using the model for all South America, are thus harder to predict. Since the latter better reflects the actual predictive conditions for which the model is going to be used (Figure \ref{fig:geodist_nocv}), kNNDM cross-validation will likely offer a more realistic map accuracy estimate \citep[see][for simulations]{Mila2022,LinnenbrinkInReview}. 

Besides the implementations of NNDM and kNNDM, CAST also offers additional spatio-temporal cross-validation strategies (\texttt{CAST::CreateSpacetimeFolds}, see \citet{Meyer2018} for an overview and examples of its use).
Note that, in general, the cross-validation strategy will not necessarily change the predictions since the model used to create the final map generally uses all samples. Therefore, if the tuning parameters (e.g. mtry in a random forest) and features stay the same, both models will result in the same prediction of species richness in South America.

%%%%%%%%%%%%%%%%%%%%%%%%%%%%%%%%%%%%%%%%%%%%%%%%%%%%%%%
\section{Spatial model tuning and feature selection}
%%%%%%%%%%%%%%%%%%%%%%%%%%%%%%%%%%%%%%%%%%%%%%%%%%%%%%%
A suitable cross-validation strategy may not only support a more reliable estimation of the map accuracy (in the absence of suitable test data) but can also improve predictions if used in the process of model selection, i.e. when choosing the predictors (feature selection) and hyperparameters (hyperparameter tuning) that yield the best performance. By accounting for a suitable spatial cross-validation strategy, the aim is to select the hyperparameters/variables that lead to the best spatial prediction for a given task. The relevance of spatial hyperparameter tuning has been shown by \citet{Schratz2019}. The process is rather straightforward: different sets of hyperparameters are tested and compared via a suitable cross-validation strategy that adequately reflects the predictive conditions of the model (e.g. kNNDM).

Using a suitable cross-validation strategy is also of high relevance during feature selection. This is especially the case if the difference between a spatial and a random cross-validation performance is large, as this may indicate significant spatial over-fitting: the model may well predict the training data but is not able to make spatial predictions beyond them. This difference is possibly due to the "clever Hans effect", a phenomenon whereby a model appears to have high predictive performance but is, in reality, relying on non-predictive features (in line with the horse "Clever Hans" that had been famous for being capable of maths but in reality was only reacting to unintentional signals of the owner). In the context of spatial prediction models, the Clever Hans effect can occur when the model learns to predict the outcome based on spurious correlations that exist in the data rather than true underlying relationships \citep[e.g.][]{lapuschkin_2019}. This can lead to inaccurate predictions when the model is applied to new areas that have not been covered by the training data. To avoid the Clever Hans effect, domain knowledge regarding which are the relevant predictors for a given outcome is key to identify a suitable set of candidate features. This prior selection can be complemented in a semi-automatic way by conducting a spatial feature selection, where we explicitly test which variables are suitable for predictive mapping and which variables aren't because they are either irrelevant or act as clever Hans predictors. 

During feature selection, we usually try to avoid testing all sets of potential predictor variable combinations for computational reasons; therefore, different feature selection strategies such as recursive feature elimination and others have been implemented in caret as well as in the other wrapper packages. Here, it is of relevance that the selection of variables is based on an appropriate cross-validation strategy that reflects the predictive conditions of the model. E.g. recursive feature elimination as implemented in caret is based on variable importance ranking from bootstrapping, a resampling method that, similar to k-fold random cross-validation, does not consider the geographical space in its default implementation. Hence, variables that lead to overfitting are likely to be top ranked and hence selected.
In CAST, we implemented a forward feature selection that first tests, for each combination of two predictor variables, which features lead to the best model, defined as the model leading to the lowest cross-validation error (i.e. kNNDM cross-validation error). Then, the number of variables is sequentially increased by choosing the predictor leading to lowest error until none of the remaining variables can further improve the performance.
The R-code to run such a forward feature selection (\texttt{CAST::ffs}) is, again, very similar to the \texttt{caret::train} function we used before.

\vspace{0.3cm}
\begin{minted}[fontsize=\small]{R}
# feature selection and final model training:
# Note that a suitable cross-validation is required as trControl
rfmodel_ffs <- CAST::ffs(training_data[,predictors],
                         training_data[,response],
                         method = "ranger",
                         importance = "permutation",
                         trControl = tr_cntrl) 
# get validation statistics and predict:
CAST::global_validation(rfmodel_ffs)
prediction <- terra::predict(rfmodel_ffs, predictors_raster)
plot(rfmodel_ffs, plotType = "selected")
\end{minted}
\vspace{0.3cm}

\begin{figure}
    \centering
    \includegraphics[width=\textwidth]{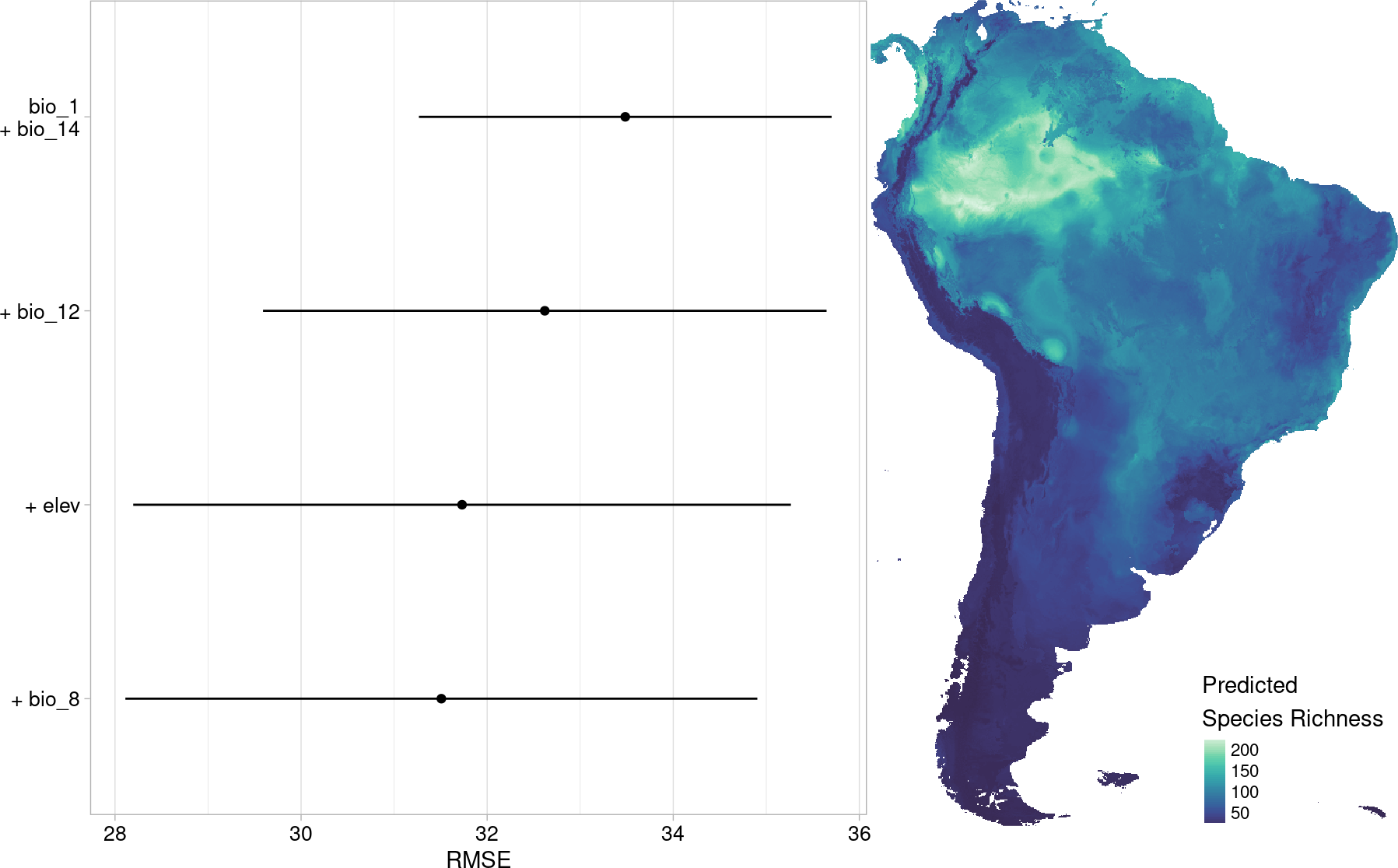}
    \caption{Selected variables that decrease the RMSE estimated with kNNDM cross-validation (left) and predicted species richness using the simplified model (right)}
    \label{fig:ffs}
\end{figure}

As we can see in Table \ref{tab:cv_results}, the model that used only a subset of the predictors (see Figure \ref{fig:ffs} for the selected predictors) led to a higher spatial mapping performance compared to the full model.

\begin{table}[!ht]
 \caption{Validation metrics resulting from a random k-fold cross-validation and kNNDM cross-validation (based on global validation). The table shows results for the full model, and for the model that was simplified using forward feature selection based on kNNDM cross-validation. For comparison, the random cross-validation performance of the simplified model (* simplified via kNNDM) is shown. A higher Root Mean Square Error (RMSE) and a smaller R$^2$ indicate a worse map accuracy.}
\centering
 \begin{tabular}{c c c c c} 
 \hline
  Model & Cross-validation & RMSE & R$^2$ & Predictors \\ 
 \hline
 Full & Random k-fold & 24.16 & 0.71 & 11\\  
 Full & kNNDM & 33.34 & 0.47 & 11\\ 
 Simplified & kNNDM & 31.97 & 0.52 & 5\\
 Simplified & Random k-fold & 24.31 & 0.71 & 5* \\
 \hline
 \end{tabular}
 \label{tab:cv_results}
\end{table}

%%%%%%%%%%%%%%%%%%%%%%%%%%%%%%%%%%%%%%%%%%%%%%%%%%%%%%%
\section{Assessment of the area of applicability}
%%%%%%%%%%%%%%%%%%%%%%%%%%%%%%%%%%%%%%%%%%%%%%%%%%%%%%%

So far, we have developed a machine learning workflow that is designed for spatial predictions by implementing a suitable cross-validation strategy for model optimization and accuracy assessment. We have also seen how, technically, predictions are possible for the entire South America by simply using the fitted model for the entire continent. However, we have not yet tested whether we should have done so. While machine learning models such as random forests can fit nonlinear relationships between predictors and response well, they are not suitable for extrapolation in the predictor space. Hence, to arrive at predictions for the entire South America, the predictor values (and their combinations) found in the prediction area need to be represented in the training data to avoid extrapolation. This, however, is rarely the case in predictive mapping applications in the field of ecology. For instance, in our example, considering the diversity of climatic conditions in the South American continent, it is unlikely that our comparably few reference samples cover all these distinct environments.

To avoid predictions to such "unknown" areas, we recently suggested the area of applicability, defined as the area for which the model was enabled to learn relationships, and where we assume that the estimated cross-validation performance holds \citep[see][for details]{Meyer2021}. The delineation of the area of applicability is based on a dissimilarity index (DI) that measures, for each prediction location, the distance to the nearest training data point in the multidimensional predictor space. The predictor space is previously normalized and (optionally) weighted by the relevance of the predictors within the trained model. Currently, Euclidean and Mahalanobis distances are implemented as distance measures.

The area of applicability is then derived by applying a threshold to the computed values of the dissimilarity index in order to identify pixels that are too dissimilar from the reference samples. In \citet{Meyer2021} we suggested that this threshold should be derived from cross-validation. If a prediction location is more dissimilar than what has been observed during cross-validation, it is considered to be outside the area of applicability. For those locations, the estimated cross-validation performance will not be considered as valid, since these dissimilar conditions have not been explored during the cross-validation.

The dissimilarity index is based on the distance to the nearest training data point in the multidimensional predictor data space, thus local training data point densities (LPD) are not considered during this process. Due to this, the index does not discriminate between areas where few, or even one isolated training data point is nearest, and a prediction location that is densely covered by training data points. As we assume that training data point densities can be highly decisive for prediction uncertainty, we suggest additionally calculating and communicating the local data point density for a prediction location (Schumacher et. al., in prep). The local data point density reflects the number of training data points that are within the threshold of the area of applicability.

The dissimilarity index, the local training data point density, and the area of applicability can be derived via the \texttt{CAST::aoa} function, and the result is an object of class "AOA". However, it is also possible to first calculate the dissimilarity index of the training data (which is the baseline to derive the threshold and hence the area of applicability) as a previous step, which is especially useful if the prediction area is large and the aoa function is applied to several tiles of the area in parallel. In this case, \texttt{CAST::trainDI} is calculated first and the result is then passed to the \texttt{CAST::aoa} function. Here, we will show the first way, i.e. directly using the \texttt{CAST::aoa} function, but a vignette on the parallelization is included in the package.

\vspace{0.3cm}
\begin{minted}[fontsize=\small]{R}
# estimate the area of applicability (AOA):
AOA <- CAST::aoa(predictors, rfmodel_ffs, LPD=TRUE)
plot(AOA)
# visualize the DI, LPD and the AOA:
plot(AOA$DI)
plot(AOA$LPD)
plot(AOA$AOA)
\end{minted}
\vspace{0.3cm}

\begin{figure}
    \centering
    \includegraphics[width=.8\textwidth]{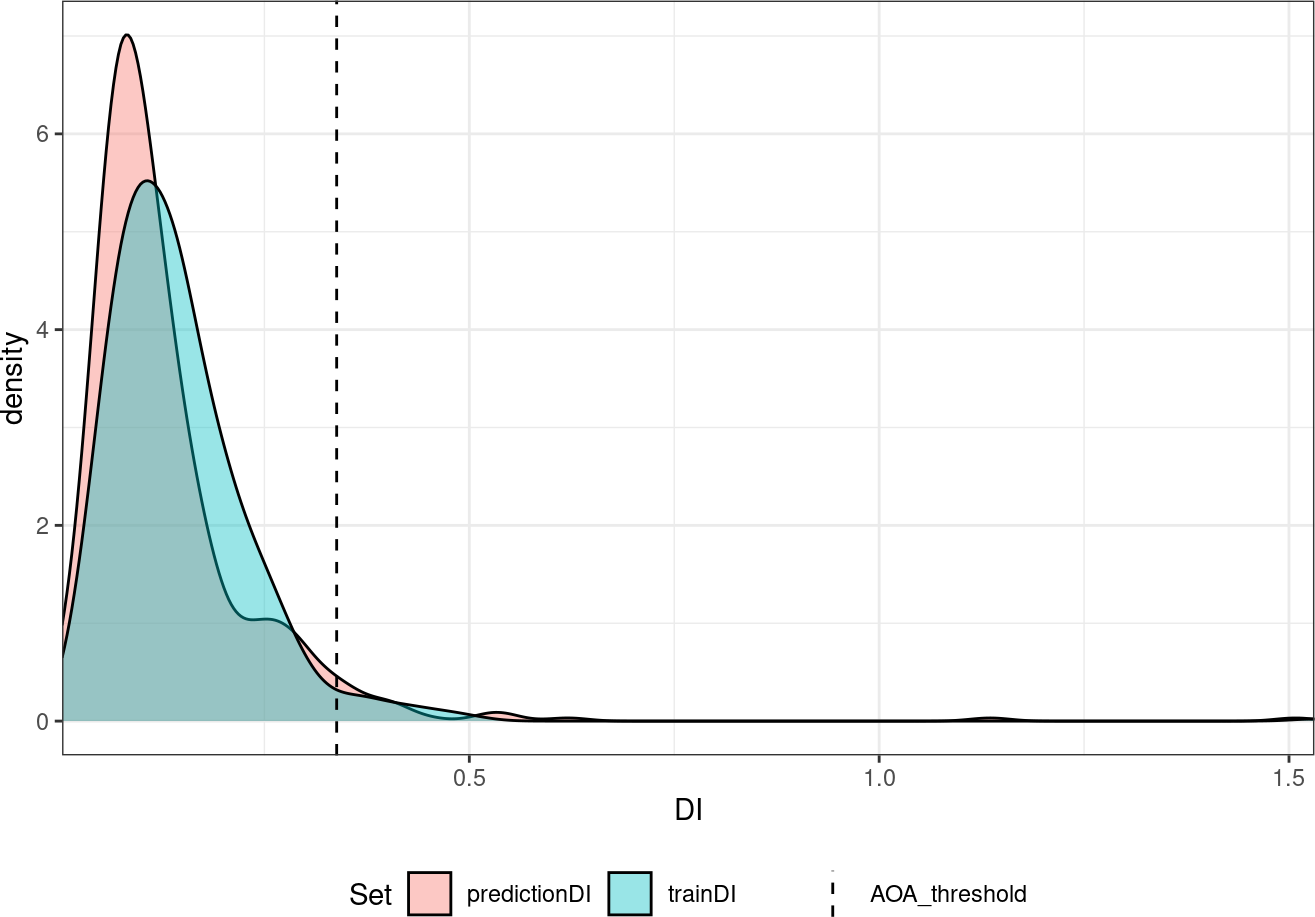}
    \caption{Output of \texttt{plot(AOA)}, Dissimilarity Index of training and prediction data with AOA threshold}
    \label{fig:aoaplot}
\end{figure}

\begin{figure}
    \centering
    \includegraphics[width=\textwidth]{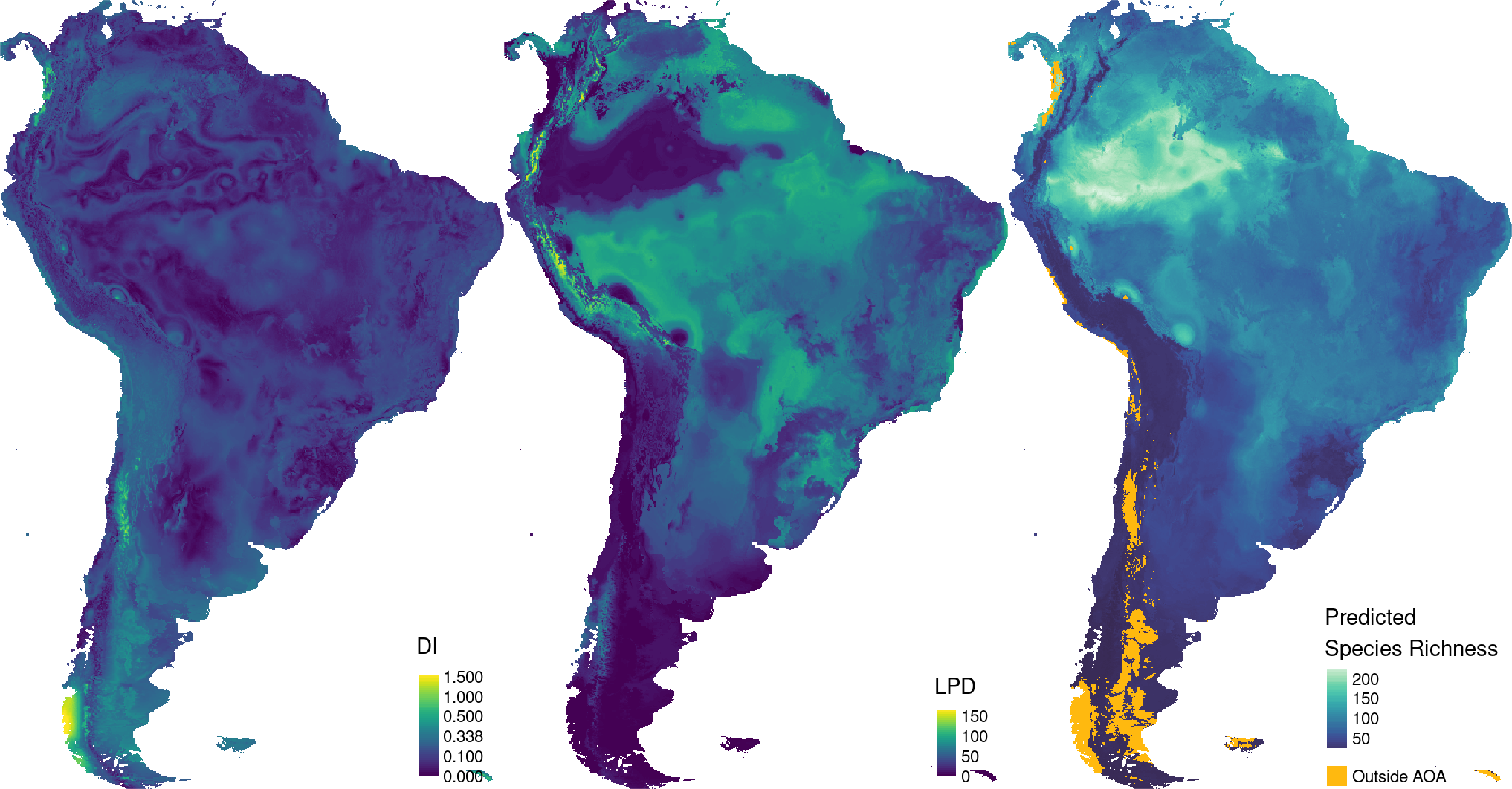}
    \caption{Dissimilarity Index (DI; left), Local Point Density (LPD; middle) and predicted species richness based on the simplified model, but only for the area of applicability (AOA; right)}
    \label{fig:ffsAOA}
\end{figure}

As we can see, some areas show high dissimilarity index values (Figure \ref{fig:ffsAOA} left) because they have very different values for the predictor variables used in the model. Based on the dissimilarity index of the training data and the model fitted in the previous chapter using kNNDM cross-validation, the threshold was derived (Figure \ref{fig:aoaplot}), which was used to mask the area for which the model is not considered to be applicable (Figure \ref{fig:ffsAOA} right).

\section{Pixel-wise performance estimation}
The dissimilarity index is not only the basis to delineate the area of applicability, but can also be used to estimate the prediction performance on a pixel-level based on the relationship between the dissimilarity index and prediction errors derived via cross-validation \citep[described in more detail in][]{Meyer2021}. This is done using \texttt{CAST::errorProfiles}. To be able to test the model performance for very low and high values of the dissimilarity index, the model may be re-fitted and validated with multiple cross-validations where the cross-validation folds are defined by clusters in the predictor space, ranging from three clusters to N clusters (equivalent to leave-one-out cross-validation). In this way, a large range of dissimilarity index values is created during the different cross-validations. Again, the errors on a pixel-level should only be estimated within the area of applicability. Note that multiple cross-validation may change the area of applicability: since multiple cross-validations will produce a large range of dissimilarities (including extrapolation situations) the threshold will likely become larger because a validation was possible for areas with higher dissimilarity. In this case, the area of applicability has to be updated accordingly by applying the new threshold to the dissimilarity index.

\vspace{0.3cm}
\begin{minted}[fontsize=\small]{R}
# calculate model:
performancemodel <- CAST::errorProfiles(rfmodel_ffs, AOA)
plot(performancemodel)
# apply model:
expected_RMSE <- terra::predict(AOA$DI, performancemodel)
# mask predictions with the AOA:
expected_RMSE = terra::mask(expected_RMSE, AOA$AOA, maskvalues = 0)

\end{minted}
\vspace{0.3cm}

\begin{figure}[ht]
    \centering
    \includegraphics[width=\textwidth]{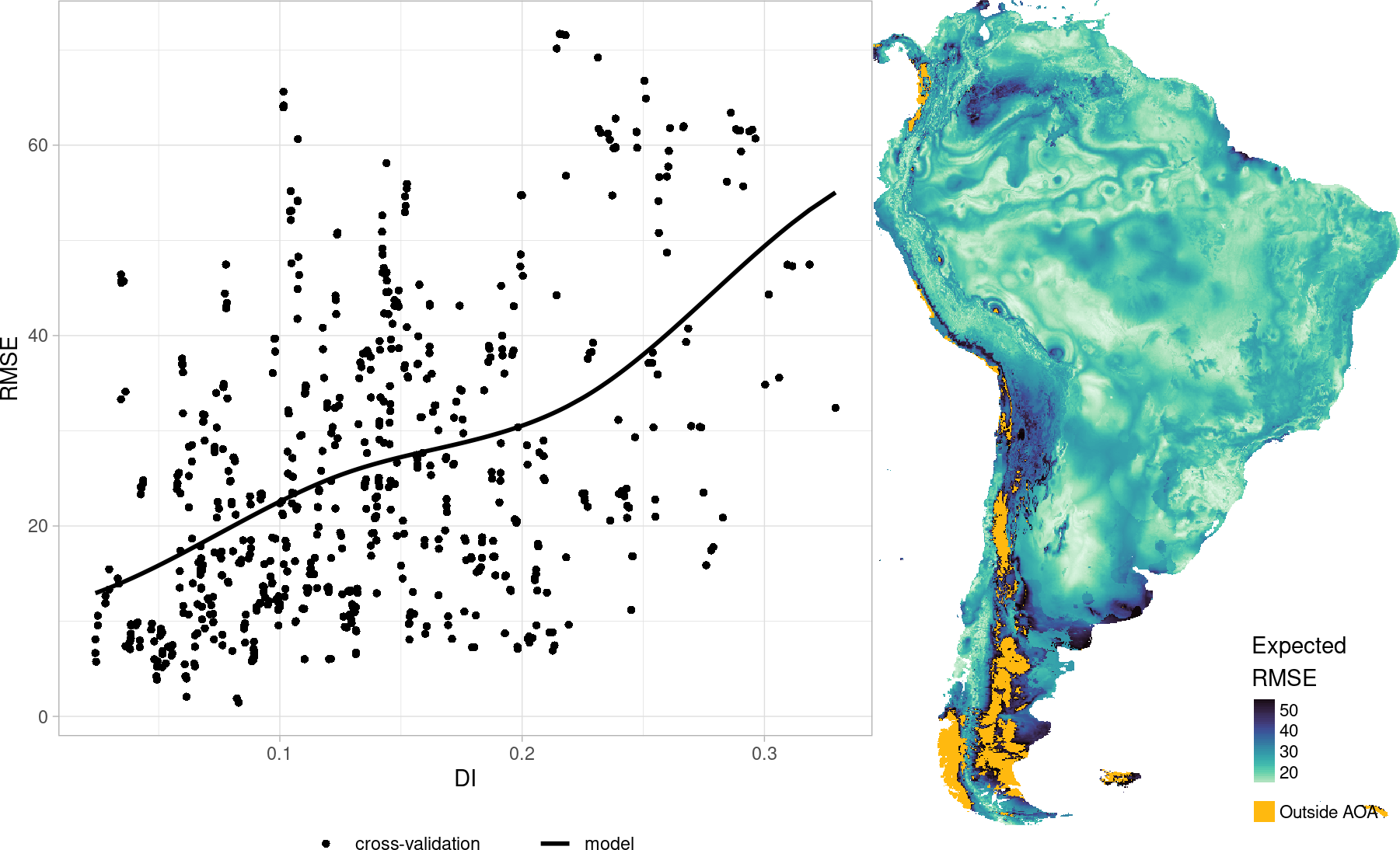}
    \caption{The modelled relationship between the dissimilarity index (DI) and cross-validation RMSE (left) and the resulting map of the expected error (right).}
    \label{fig:ffsExpectedModel}
\end{figure}

The resulting object contains the model representing the relationship between the dissimilarity index and the performance metric. It can be used to predict the error (or performance) on a pixel level. The results for our example are shown in Figure \ref{fig:ffsExpectedModel} left: The performance (per default the same that was already used during model tuning; here: RMSE) is calculated for moving windows of dissimilarity index values based on the cross-validated training data. A shape-constrained additive model is then used to model the relationship between the performance metric and the dissimilarity values \citep[see][for details]{Meyer2021}. The model is, in a second step, applied to each prediction location (i.e. pixel) to map the expected performance (Figure \ref{fig:ffsExpectedModel} right). Analogous to the workflow described above, it is also possible to use the local data point density for a pixel-wise performance estimation.
We have, however, to keep in mind that the estimated performance is based here on the dissimilarity (or local data point density) in predictor space only. Hence, other sources of uncertainty are not reflected and may be accounted for by other methods (e.g. quantile random forest).

%%%%%%%%%%%%%%%%%%%%%%%%%%%%%%%%%%%%%%%%%%%%%%%%%%%%%%%
\section{Conclusion and outlook}
%%%%%%%%%%%%%%%%%%%%%%%%%%%%%%%%%%%%%%%%%%%%%%%%%%%%%%%
Here, we presented the R package CAST to support the process of spatial model training and evaluation. We first introduced kNNDM cross-validation to provide more realistic estimates of map accuracy, and showed how it can be used in a feature selection procedure to reduce the risk of spatial overfitting and eventually improve the spatial model performance. Next, the area of applicability was presented with the aim of limiting predictions to the areas where the model was enabled to learn about relationships. Finally, we showed how the dissimilarity index based on distances in the predictor space, or the local training point density, can be used to quantify uncertainties that originate from different predictor properties than those of the training data.

The CAST package is still under active development and we intend to extend the functionalities. Future developments will include further model inspection tools, (k)NNDM in predictor space, as well as a stronger support for integration into mlr3 and tidymodels workflows.

%%===========================================================================================%%

\bibliographystyle{agsm}
\bibliography{sn-bibliography}

\end{document}